# Characterizing player's playing styles based on Player Vectors for each playing position in the Chinese Football Super League

Yuesen Li[1], Shouxin Zong[1], Yanfei Shen[1*], Zhiqiang Pu[2], Miguel-Ángel Gómez[3], Yixiong Cui[1,4*]

1. School of Sports Engineering, Beijing Sport University, Beijing, China.
2. Institute of Automation, Chinese Academy of Sciences, Beijing 100190, China
3. Facultad de Actividad Física y del Deporte (INEF), Universidad Politécnica de Madrid, Madrid, Spain.
4. AI Sports Engineering Lab, School of Sports Engineering, Beijing Sport University, Beijing, China.

**Correspondence:** Yixiong Cui, Yanfei Shen; *School of Sports Engineering, Beijing Sport University, Information Road 48, Haidian District, 100084, Beijing, People's Republic of China*. E-mail: (amtf000cui@gmail.com; syf@bsu.edu.cn).

# Abstract

Characterizing playing style is important for football clubs on scouting, monitoring and match preparation. Previous studies considered a player's style as a combination of technical performances, failing to consider the spatial information. Therefore, this study aimed to characterize the playing styles of each playing position in the Chinese Football Super League (CSL) matches, integrating a recently adopted *Player Vectors* framework. Data of 960 matches from 2016-2019 CSL were used. Match ratings, and ten types of match events with the corresponding coordinates for all the lineup players whose on-pitch time exceeded 45 minutes were extracted. Players were first clustered into 8 positions. A player vector was constructed for each player in each match based on the *Player Vectors* using Nonnegative Matrix Factorization (NMF). Another NMF process was run on the player vectors to extract different types of playing styles. The resulting player vectors discovered 18 different playing styles in the CSL. Six performance indicators of each style were investigated to observe their contributions. In general, the playing styles of forwards and midfielders are in line with football performance evolution trends, while the styles of defenders should be reconsidered. Multifunctional playing styles were also found in high rated CSL players.

**Key words:** *soccer; match analysis; style of play; performance profile; machine learning*

# 1. Introduction

The research of performance analysis in association football covers various aspects such as the identification of key indicators, movement patterns or passing networks (Lord et al., 2020). Among the wide scope of this research field, the evaluation of teams' and players' match performance has recently become one of the essential areas that attract scientific attention to improve the body of research (Goes et al., 2021; Goes et al., 2020; Gudmundsson & Horton, 2017; Pappalardo et al., 2019). Relevant findings and models could not only assist scouting experts and team stakeholders when recruiting players, but also help coaches and managers to monitor the evolution of tactical-technical styles in order to adjust match strategies and control for match demands (Decroos & Davis, 2020; Zhou et al., 2021).

One of the most commonly used and developed evaluation tool is the rating and ranking system of teams and players, which could be applied to assess the strength of a team or a player objectively to a certain degree (Lasek et al., 2013). Technical and tactical data-driven rating methods such as computing a weighted sum of subindices (McHale et al., 2012), learning weights out of performance features (Pappalardo et al., 2019, Li et al., 2020) and assessing behaviors by estimating probabilities (Decroos et al., 2019) have all been recently proposed and studied. Apart from rating, characterizing playing styles is also an essential perspective of evaluating, which would provide more comprehensive insights into teams' match-play and players' actual roles, and help teams to better recruit and avoid mismatches in the transfer market.

From a team level approach, a playing style can be defined as the interactions of playing patterns (Hewitt et al., 2016) and researchers tends to combine certain attacking and defending attributes, transitions and set pieces to characterize the style of play for the purpose of winning (McLean et al., 2017). In particular, Fernandez-Navarro et al. (2016) used factor analysis via principal components analysis (PCA) to define attacking and defending styles in elite soccer leagues of Spanish La Liga and English Premier League. Following the insight and the statistical approach of these studies, Lago-Peñas et al. (2018) identified five different kinds of playing styles in Chinese Soccer Super

League: possession, set piece attack, counter attacking play and two kinds of transitional play; while Gómez et al. (2018) compared the playing styles within different team qualities and match locations based on the data of Greek professional soccer and claimed that ranking and match location are also important when identifying teams' playing styles; and Castellano and Pic (2019) combined playing styles with match outcomes in La Liga and found that well-designed offensive styles and defensive minds are keys to win.

However, research on data-driven playing style characteristic at an individual level is relatively scarce. Some studies have attempted to explore the player similarity in match-play. For example, Peña and Navarro (2015) stipulated several passing sequences and tried to find the replacement of the legendary midfielder "Xavi Hernandez" using dimensionality reduction and unsupervised clustering on different passing patterns. In addition, Mazurek (2018) searched for the most similar player to Lionel Messi, using 17 advanced match indicators most relevant to an attacking player. Moreover, Garcia-Aliaga et al. (2020) attempted to analyse patterns of play and key performance indicators by position using machine learning methods and verified the possibility of characterizing players' positions using only game statistics. Although these studies provided some preliminary findings, their practical usefulness were limited due to the fact that a player's style was only considered as a combination of technical performances, failing to control for the spatial distribution of these actions.

In light of this phenomena, Decroos and Davis (2020) proposed a *Player Vectors* model to express football player's playing style based on the concept that a player's style can be expressed by his/her preferred areas on the field and which action he or she tends to perform in each of these locations. Using a framework with Non-negative Matrix Factorization (NMF) as the core, they extracted components from heatmaps, representing different kinds of playing styles and the similarities of a player's style with these components being constructed to the final player vectors, thereby characterizing playing styles.

Players in different roles have different responsibilities, previous research have explored and detected player roles based on their active positions (Bialkowski et al.,

2014; Pappalardo et al., 2019). However, it should be noted that there is a wide variety of playing styles in these positions, and they should play as different roles. Previous research (Aalbers & Van Haaren, 2019) has classified players European top leagues into various playing styles using machine learning models, but the players were initially labeled by experts before the modeling, which might limit the applicability to players of other tournaments. Therefore, this study aimed to provide a data and domain knowledge driven solution to characterize players' playing styles in different positions in professional football, integrating the state-of-the-art method *Player Vectors*. Such approach will allow to analyze the differences among playing styles of players in the CSL and compare these playing styles with the current tactical requirements for players in professional football.

## 2. Methods

### *2.1 Sample and data source*

Data related to match events and player information of all 960 matches from 2016-2019 CSL where 22 teams and 912 players competed were obtained from a publicly-assessed football statistics website "whoscored.com", whose original data are provided by OptaSports Company. The validity of their data collection has been previously validated (Liu et al., 2013).

The final match event data-set contains 44 types of actions related to shooting, organizing, skill, defending and goalkeeping. Each action has its own result and attribute (See Supplementary Table 1 for detailed definitions).

### *2.2 Player Vectors*

Fixed-size player vectors that characterizes players' playing style were built followed by the framework showed in Figure 1.

The framework contains three major parts: *Constructing*, *Compressing* and *Assembling*. The Constructing part draws normalized smooth heatmaps with the size of $m \times n$ for each player of action $t$. Then, the Compressing part applied a non-negative matrix factorization (NMF) process to compress the matrix $M = [x_0, x_1, \ldots, x_l]$ which columns are the vectorized 1-dimensional heatmap vector $x$ with the length of $mn$ for each player.

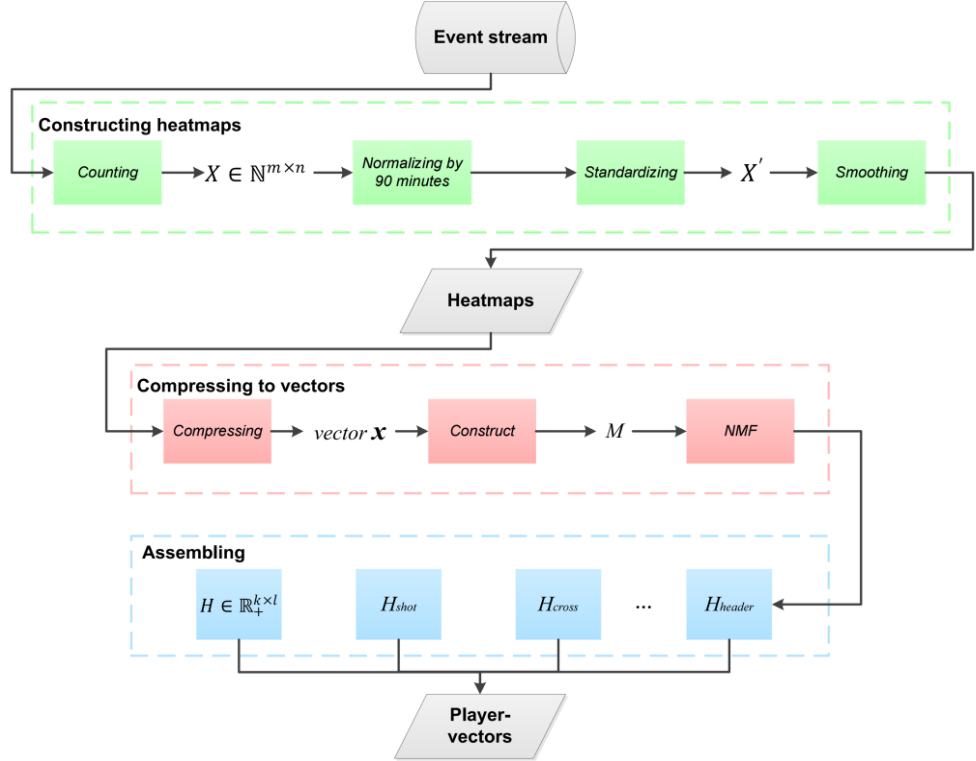

**Figure 1.** Flowchart of the *Player Vectors* framework

NMF is a form of PCA where the resulting components contain only positive values (Lee & Seung, 2000). This results in two matrices $W$ and $H$ such that:
$$M \approx WH$$
Where $M \in \mathbb{R}_+^{mn \times l}$, $W \in \mathbb{R}_+^{mn \times k}$, $H \in \mathbb{R}_+^{k \times l}$. $k$ is a user-defined parameter that refers to the number of principal components for action type $t$. The columns in $W$ are the principal components that represent $k$ feature heatmaps. The rows in $H$ are vectors that are the compressed heatmaps in $M$ which can be seen as the player playing style's similarities to each feature heatmaps. The actions used in this study is showed in Table 1.

**Table 1.** Actions used in the *Player Vectors* framework.

| Actions | Definitions |
| --- | --- |
| Shot | An attempt to score a goal, made with any (legal) part of the body, either on or off target |
| Cross | Any intentional played ball from a wide position intending to reach a teammate in a specific area in front of the goal. |
| Dribble | Player dribbles at least 3 meters with the ball. |
| Pass | Any intentional played ball from one player to another. |
| Long pass | A lofted ball where there is a clear intended recipient, must be over shoulder height and using the passes height to avoid opposition players or a long high ball into space or into an area for players to chase or challenge for the ball |
| Key pass | The final pass or pass-cum-shot leading to the recipient of the ball having an attempt at goal. |
| Interception | Where a player reads an opponent's pass and intercepts the ball by moving into the line of the intended pass. |
| Clearance | A defensive action where a player kicks the ball away from his own goal with no intended recipient. |
| Header | Any action using a player's head, whether it is a pass or a dual in the air. |
| Recovery | Where a player recovers the ball in a situation where neither team has possession or where the ball has been played directly to him by an opponent, thus securing possession for their team. |

The assembling part concatenates the $H$ for each action type to $H' \in \mathbb{R}_+^{(k_1+k_2+\cdots+k_t) \times l}$ and the player vector $v$ of a player $p$ is the corresponding row in $H'$.

Finally, the similarity of the playing style of each player was calculated as the Manhattan distance between the player vectors. To illustrate the similarity by percentage, it was further transformed by:

$$S = \frac{(D_{ij} - D_{max})}{D_{max}} \times 100\%$$

where $D_{ij}$ is the Manhattan distance between player $i$ and $j$, $D_{max}$ is the maximize distance between each pair of players in data-set.

## 2.3 Position clustering

A player's position can be changed in different matches (Pappalardo et al., 2019) and the actual position is sometimes varied to which given in the starting lineups. To eliminate the resulting potential impacts as much as possible, it is necessary to detect the real position of the players.

A total 18,784 of 19,200 average positions of the actions of every starting player who played more than 45 minutes in each match were extracted from the data-set. A *k*-means algorithm was run on those average positions. A silhouette score of each sample was used to find out the most proper k value among 5 to 10, which is computed as:

$$s(i) = \frac{b(i) - a(i)}{\max\{a(i), b(i)\}}$$

where $a(i)$ is the average distance between sample *i* to the other samples in the cluster and $b(i)$ is the distance between *i* and its nearest cluster (Pedregosa et al., 2011; Rousseeuw, 1987).

In football matches, side switching is a classical tactic which will lead to a cluster error for the wide players (being clustered to the center of the field). After finding only 87 times of side switching, these samples were deleted and led the final clustering samples to 18,699. Further partition of the positions and style clustering were based on the result of this position clustering process.

## 2.4 Style clustering

To extract different types of playing styles and the similarity between an outfield player's style of play and these representative styles, another *n* NMF processes were carried out for the *n* different positions respectively using different actions. These were acted on the starting lineups in each single match with goalkeepers excluded. The most similar style was considered as the playing style of an individual player in a specific match. Each playing style was defined based on their different preferences observed from the Player Vectors and referred to a well-known sport video game—Football

Manager. Total appeared numbers (N), Appeared numbers of domestic players (ND) and foreign players (NF), Win-Loss ratio (W/L), match ratings (R) and 3 performance indicators (Goals, Shots and Assists) of each style were calculated to compare their contributions. The numbers of the official starting position which were observed more than 100 times in each style were also extracted (See Supplementary Table 2). Those starting positions are Forward (FW), Midfielder (MF), Center Defender (CD), Left/Right Midfielder (L/RM) and Left/Right Defender (L/RD).

## 2.5 Player Comparison

Comparing a player's playing style before and after a specific scenario is valuable and helpful on evaluating a player's performance and a club's strategic decision on player transfer and recruitment. This study compared three pairs of player's reports representing three different scenarios:

**(i)** **The most similar players:** Alex Teixeira (Jiangsu Suning) vs. Wei Shihao (Guangzhou Evergrande). Wei is the most similar player of Teixeira according to the output of the model;

**(ii)** **Player progressing:** Wu lei (Shanghai SIPG) in season 2017 vs. Wu Lei in season 2018. Wu is considered to be one of the best Chinese players and won the Player of the season in 2018. Such comparison would help explore how his match play progressed;

**(iii)** **Midfield partners:** Renato Augusto (Beijing Guoan) vs. Jonathan Viera (Beijing Guoan). Being the two of only four players who performed double figures for both goals and assists in season 2018 (Whoscored, 2018), these two midfielders were clearly indispensable for Beijing Guoan. Comparing their differences in playing style might offer a better understanding of their tactical differences on the field.

Each player report contains five parts: (i) Player basic information; (ii) the Similarity between them; (iii) Stats of the player's styles in each match; (iv) the Figure of the stats; and (v) the General player vector of the season.

# 3. Results

## 3.1 Player Vectors

The NMF process generated a player vector of 44 dimensions for each player, the component heatmaps are showed in Figure 2.

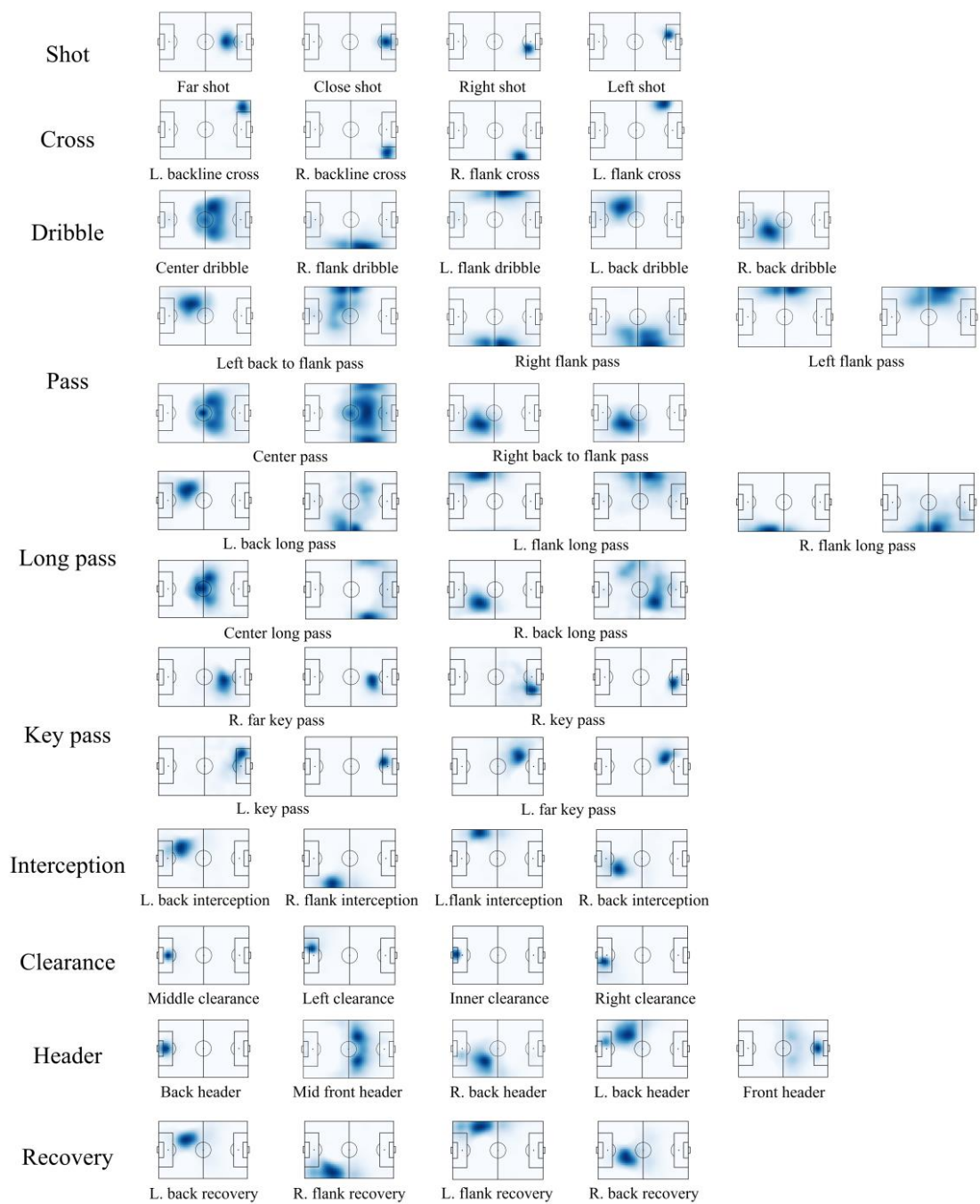

**Figure 2.** The heatmaps of each component

## 3.2 Position clustering

Figure 3 (a) shows the position clusters resulted by the k-means algorithm with a most proper *k*=8, which has a stable silhouette score (ss=0.41) among 10 times of experiments with random sets of centroids that were used to initiate the algorithm. The clusters located in the two sides were merged into one position manually which led to the final positions reduced to five: Strikers (ST), Left/Right Wing forwards (L/RW), Central Midfielders (CM), Left/Right Full Backs (L/RFB), and Central backs (CB). The actions used in different positions in the following style clustering process were also showed in Figure 3 (b).

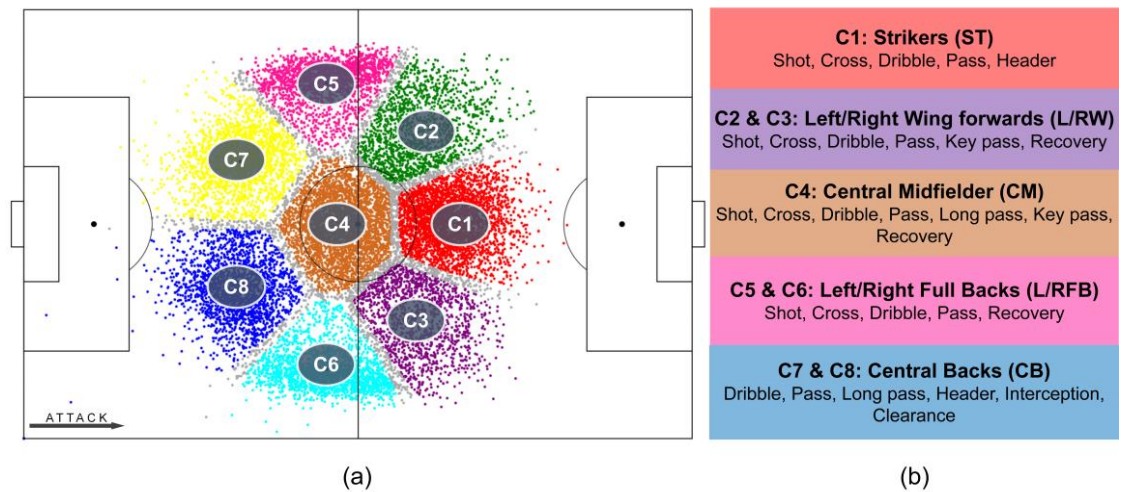

**Figure 3.** Eight positions clustered by the K-means algorithm and the actions used in different positions in style clustering

## 3.3 Style clustering

Table 2 shows the style clustering and descriptive (mean and standard deviation, SD) results for each style. The definition of each style is illustrated in Supplementary Table 2.

**Table 2.** Style clustering and descriptive statistic results of each style.

| Position | Style | Appearances | | | Rating (SD) | Goals (SD) | Shots (SD) | Assists (SD) | Win/Loss |
|---|---|---|---|---|---|---|---|---|---|
| | | Total | Domestic | Foreign | | | | | |
| ST | Second Striker | 1394 | 340 | 1054 | 7.22 (.95) | 0.36 (.59) | 2.86 (2.05) | 0.21 (.47) | 1.01 |
| | Target Man | 570 | 123 | 447 | 7.07 (.82) | 0.34 (.56) | 2.23 (1.49) | 0.10 (.31) | 0.88 |
| | Mobile Striker | 646 | 243 | 403 | 7.07 (.88) | 0.33 (.59) | 2.22 (1.64) | 0.15 (.39) | 1.12 |
| | Poacher | 613 | 154 | 459 | 7.38 (1.01) | 0.65 (.80) | 4.23 (2.01) | 0.12 (.34) | 1.05 |
| CM | R Defensive Midfielder | 760 | 595 | 165 | 6.70 (.59) | 0.05 (.22) | 0.80 (1.03) | 0.05 (.22) | 1.04 |
| | Playmaker | 1549 | 1062 | 487 | 6.83 (.69) | 0.09 (.32) | 1.14 (1.21) | 0.08 (.29) | 0.83 |
| | L Defensive Midfielder | 803 | 595 | 208 | 6.78 (.64) | 0.09 (.31) | 0.77 (1.01) | 0.07 (.26) | 0.98 |
| | Wide Midfielder | 544 | 373 | 171 | 6.85 (.75) | 0.14 (.39) | 1.17 (1.20) | 0.15 (.38) | 1.18 |
| L/RW | L Winger | 1307 | 918 | 389 | 6.92 (.80) | 0.19 (.46) | 1.59 (1.48) | 0.16 (.42) | 1.08 |
| | R Winger | 1224 | 928 | 296 | 6.85 (.75) | 0.18 (.42) | 1.46 (1.41) | 0.14 (.38) | 1.04 |
| | Inside Forward | 890 | 450 | 440 | 7.01 (.87) | 0.24 (.53) | 2.03 (1.77) | 0.14 (.40) | 1.02 |
| L/RFB | R Wing Back | 1639 | 1609 | 30 | 6.65 (.59) | 0.02 (.13) | 0.35 (.63) | 0.06 (.25) | 1.01 |
| | L Wing Back | 1613 | 1603 | 10 | 6.70 (.57) | 0.01 (.12) | 0.40 (.68) | 0.07 (.27) | 1.03 |
| | L Back | 347 | 306 | 41 | 6.64 (.54) | 0.04 (.21) | 0.51 (.82) | 0.05 (.21) | 0.83 |
| | R Back | 386 | 358 | 28 | 6.54 (.59) | 0.03 (.17) | 0.38 (.66) | 0.04 (.20) | 0.91 |
| CB | R Ball Playing Defender | 1750 | 1410 | 340 | 6.60 (.59) | 0.03 (.18) | 0.31 (.60) | 0.01 (.12) | 0.98 |
| | L Ball Playing Defender | 1979 | 1406 | 573 | 6.68 (.61) | 0.04 (.21) | 0.39 (.68) | 0.02 (.13) | 0.96 |
| | Central Defender | 685 | 510 | 175 | 6.80 (.63) | 0.03 (.17) | 0.30 (.59) | 0.02 (.15) | 1.66 |

Notes: **ST** = Strikers, **L/RW** = Left/Right Wing forwards, **CM** = Central Midfielders, **L/RFB** = Left/Right Full Backs, **CB** = Central Backs

**Strikers**

As it was showed in Figure 4 (a), The CSL Strikers were classified into four playing styles: Poacher (preference for close shot and front header), Second Striker (preference for Long Shot, Center Dribble and Center Pass), Mobile Striker (high preference for L/R shot, L/R backline cross, L/R flank dribble and L/R flank pass) and Target man (high preference for Mid front header). Among a total of 3,223 observations, the best rated style was Poacher with the highest goals per match, followed by Second Striker and Mobile Striker. On the other hand, Target Mans had the lowest Win/Loss ratio among the four styles. A large amount of foreign MFs were found to be Second Striker and Mobile Striker.

**Central Midfielders**

The Central Midfielders were clustered into four styles (see Figure 2 (b)): Playmaker (preference for Dribble, Pass and Long pass in the middle of the field and no preference for the defensive actions), L and R Defensive midfielders (preference for Dribble, Pass, Long pass and Ball recovery in the back of the pitch) and Wide midfielder (preference for Flank dribble, Flank pass, Flank long pass, Ball recovery and Aerial dual in the mid front area). The best rated style among a total 3,656 observations was Wide Midfielder who performed the best in goals and assist as well as the Win-Loss ratio. The second-best rated style was Playmaker while its Win-Loss ratio was the lowest. R and L defensive Midfielders' Win-Loss ratio was relatively balanced. The starting positions with more than 100 observations in all these styles are solely MF.

**Left/Right Wing forwards**

A total number of 3,421 L and R Wing Forwards were clustered into three styles which are showed in Figure 2 (c): Inside Forward (preference for Center dribble and pass, Close shot and Long shot) and L/R Winger (preference for Cross in the flanks and backline and dribble, pass in the flanks). Among them, Inside Forward got the highest rating. Most of the L/R Wingers were domestic players, while foreign FWs and domestic MFs were the majority of Inside Forwards.

**Left/Right Full Backs**

As it was illustrated in Figure 2 (d), 3985 L and R Full Backs were classified into four playing styles: L/R Wing Backs (preference for Flank dribbles, passes and Flank, Backline crosses) and L/R Backs (preference for Back dribbles, passes and recoveries). L/R Wing Backs performed better both in rating and Win-Loss ratio than L/R Backs. Almost all Full Backs were domestic and played on their original starting position (L/RD).

**Central backs**

Central backs were clustered into three different styles (see Figure 2 (e)): L and R Ball Playing Defender (preference for Back dribbles, passes and long passes) and Central Defender (preference for Back header and clearances in all four areas and little preference for dribbles and passes). Among a total 4,414 observations, Central Defender had the best rating and Win-Loss ratio. DF was the main part of these three playing styles where most of the players were domestic.

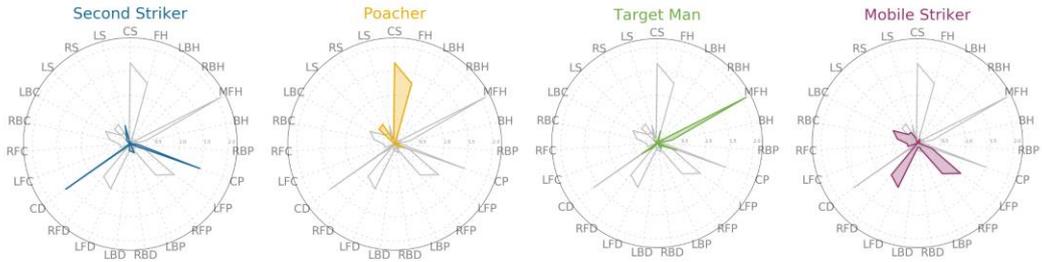

(a). Strikers

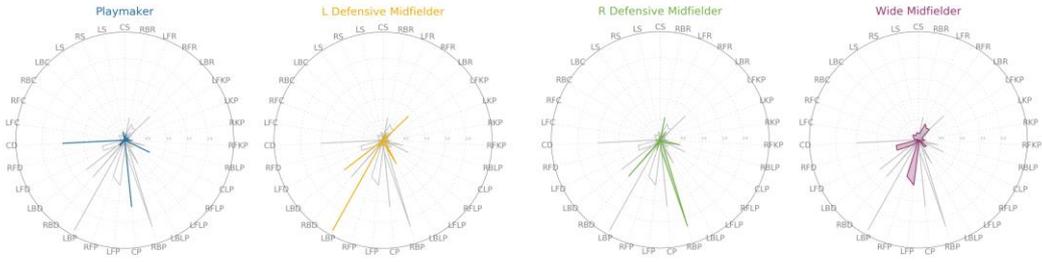

(b). Central Midfielders

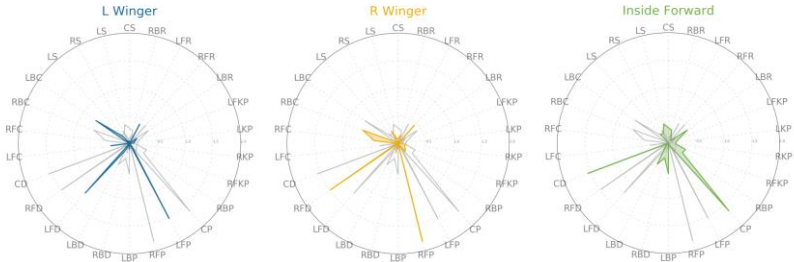

(c). Left/Right Wing forwards

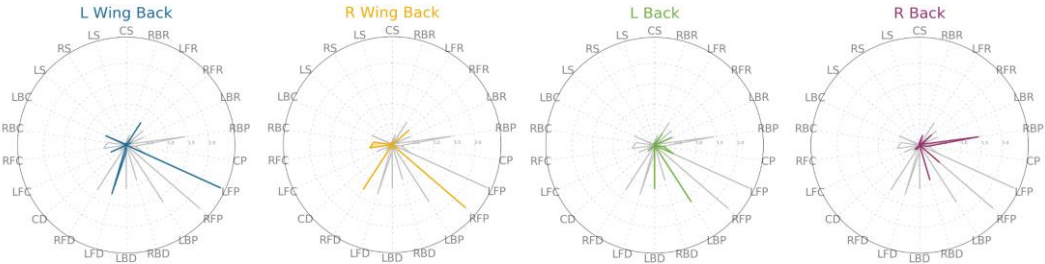

(d). Left/Right Full Backs

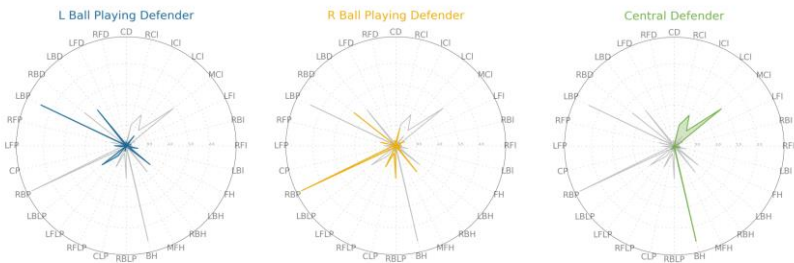

(e). Central Backs

**Figure 4.** Playing styles of each position. (Notes: **CS** = Close Shot, **LS** = Long Shot, **RS =** Right Shot, **LS** = Left Shot, **LBC** = L. Backline Cross, **RBC** = R. Backline Cross, **RFC** = R. Flank Cross, **LFC** = L. Flank Cross, **CD** = Center Dribble, **RFD** = R. Flank Dribble, **LFD** = L. Flank

Dribble, **LBD** = L. Back Dribble, **RBD** = R. Back Dribble, **LBP** = L. Back Pass, **RFP** = R. Flank Pass, **LFP** = L. Flank Pass, **CP** = Center Pass, **RBP** = R. Back Pass, **LBLP** = L. back long pass, **LFLP** = L. flank long pass, **RFLP** = R. flank long pass, **CLP** = Center long pass, **RBLP** = R. back long pass, **RFKP** = R. far key pass, **RKP** = R. key pass, **LKP** = L. key pass, **LFKP** = L. far key pass, **LBI** = L. back interception, **RFI** = R. flank interception, **LFI** = L. flank interception, **RBI** = R. back interception, **MC** = Middle clearance, **LC** = Left clearance, **IC** = Inner clearance, **RC** = Right clearance, **BH** = Back Header, **MFH** = Mid Front Header, **RBH** = R. Back Header, **LBH** = L. Back Header, **FH** = Front Header, **LBR** = L. back recovery, **RFR** = R. flank recovery, **LFR** = L. flank recovery, **RBR** = R. back recovery)

## 3.4 Player comparisons

**Alex Teixeira VS. Wei Shihao**

Figure 5 (a) shows the comparison report of Teixeira and Wei Shihao, whose similarity was 93.6%. Clearly the most appeared and rated style of Teixeira was Second Striker of ST, with an appearance of 13 times and a high average rating of 7.62 points per game. The second most played style of Teixeira was L Winger, which was the most appeared style of Wei Shihao, with a rating of 7.18. As for the season general style, Teixeira had a much higher preference for Long shot, Center dribble, Back dribbles and Key passes while Wei Shihao preferred more to cross from the right backline.

**Wu Lei 2017 VS. Wu Lei 2018**

Figure 5 (b) shows the player report of Wu Lei in both 2017 and 2018 which had a similarity of 92.8%. It is obviously that the most preferred style of Wu Lei transformed from Winger to Poacher although the average ratings of the flank roles (L Winger, R Winger and Inside Forward) were higher than that of playing as a Poacher. For the season general comparison, in season 2018 Wu Lei showed a much higher preference for Close Shot, Mid front Header and Front header; while the preference for Left shot, Crosses, Back Passes and Dribbles, all dropped.

**Renato Augusto VS. Jonathan Viera**

As it is showed in Figure 5 (c), the midfield partner had a similarity of 70.6%. Among all the 24 matches they started together, Viera played most as a Second Striker while Augusto preferred more for playing as a Playmaker or playing in the left of the field as an Inside Forward or a Winger. For the general style of the season, Viera showed a much higher preference for Close shot, Right shot, Center dribble (even though the similarity of Augusto was also relatively high) and Key passes, while Augusto did more on Long shot, Crosses, Flank dribbles, Back passes and Recoveries.

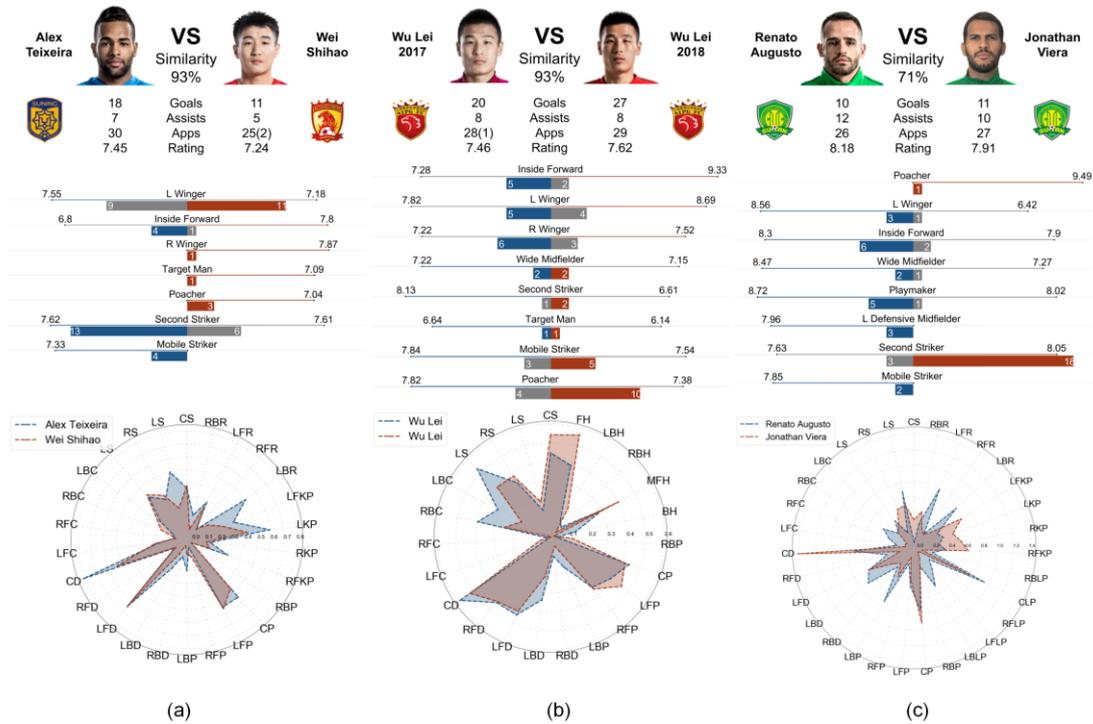

**Figure 5.** Comparison results of the three pairs of players. (Notes: **CS** = Close Shot, **LS** = Long Shot, **RS =** Right Shot, **LS** = Left Shot, **LBC** = L. Backline Cross, **RBC** = R. Backline Cross, **RFC** = R. Flank Cross, **LFC** = L. Flank Cross, **CD** = Center Dribble, **RFD** = R. Flank Dribble, **LFD** = L. Flank Dribble, **LBD** = L. Back Dribble, **RBD** = R. Back Dribble, **LBP** = L. Back Pass, **RFP** = R. Flank Pass, **LFP** = L. Flank Pass, **CP** = Center Pass, **RBP** = R. Back Pass, **LBLP** = L. back long pass, **LFLP** = L. flank long pass, **RFLP** = R. flank long pass, **CLP** = Center long pass, **RBLP** = R. back long pass, **RFKP** = R. far key pass, **RKP** = R. key pass, **LKP** = L. key pass, **LFKP** = L. far key pass, **BH** = Back Header, **MFH** = Mid Front Header, **RBH** = R. Back Header, **LBH** = L. Back Header, **FH** = Front Header, **LBR** = L. back recovery, **RFR** = R. flank recovery, **LFR** = L. flank recovery, **RBR** = R. back recovery)

## 4. Discussion

Using match event data, spatial-temporal information of actions and domain knowledge, the study was to provide a more complex solution to characterize professional football players' playing styles of different positions, integrating a recently adopted *Player Vectors* framework. Twelve on-the-ball actions were used to generate a 44-dimensional player vector for each player in each match. Players were clustered into eight different positions, and after performing another NMF process on various components of these positions, eighteen different kinds of playing styles were finally discovered in the China Football Super League. The findings provided additional insights into player evaluation from the perspective of playing style, and could be applied to player recruitment and match preparation.

## *4.1 Position Clustering*

The Position Clustering process detected eight different roles based on their average on-the-ball action positions. A silhouette score of 0.41 seemed not so high in data science resulted by the denseness of the sample, while the algorithm was able to divide the players into clusters in line with the common knowledge of soccer. Additionally, the lacking of tracking data also limited the possibility of using more complex methods which may enhance the clustering effect (Bialkowski et al., 2014). For these reasons, despite the imperfections of this position clustering method, it can meet the needs of this study.

## *4.2 Playing styles of the CSL players*

***Strikers***

This study revealed that almost half of the Center Forwards (Second Striker) in CSL tended to move deep into the midfield to help teams' collective organization (Duarte et al., 2012). However, unlike the Second Strikers, Poachers were required to focus on the most important task in football: putting the ball in the back of the net (Guide to FM, 2020), which explained their highest average rating. Moreover, Mobile Strikers with the highest W/L, who most of them were foreign players started as a striker or midfielder, were inclined to play on the flanks and to cut into the middle, implying that assigning best high-level foreign strikers to play on the flank may be a potential key to goal-scoring and success (Lago-Peñas et al., 2018). Furthermore, Target Mans had the lowest rating and W/L, which was in line with their duties: being green leaves to open up spaces for key players by their physicality and aerial presence and join in the battle in the midfield when the teams were facing unfavorable match situation (Guide to FM, 2020).

It is interesting to find out that a large number of strikers are Second Strikers. Part of the reasons may be related to the strategical demands of the match, since nearly two thirds of CSL strikers are foreign players who make indispensable contribution to teams' organizing, thus enhancing teams' success (Gai et al., 2019). By the other hand, nearly 40% of Second Strikers were players starting as a midfielder and were clustered to Strikers, and 65% of them were also foreign players (e.g., Jonathan Viera). This indicated that the CSL attacking midfielders not only participated into organizing, but also moved frequently forward to the attacking third to create threat to the goal.

### *Left/Right Wing Forwards*

L/R Wingers were the team's sharp knife, they may choose to dribble alongside the flank line and make backline crosses or attempt to threat the keeper from the wide side of the box. Inside Forward was a role similar to the Second Striker and the Playmaker but acting on a wider side of the field, which can explain the high proportion of players who started as a midfielder in this style. This playing style are space orchestrators and have become more and more necessary in modern soccer tactical, especially in an era when super full backs sometimes monopolize the team's attacking and defending task on the flank (Crossbarhub, 2020).

It is worthwhile to notice that the appearance number of L/R Wingers were nearly three times than that of Inside Forwards. The finding echoed with the study by Gai et al. (2019) that wing players in CSL were prone to play more crosses, while such behavior was only present in wingers of weak teams in Spanish La Liga (Liu et al., 2016). This may suggest that most of the CSL Wing Forwards lack the abilities to cut inside and make direct threats to the opponent goal, which would in turn restrict coaches from adopting other tactical alternatives. For the Inside Forwards, it is interesting to find out the majority of them were domestic players who started as a midfielder or foreign strikers. This might suggest that in the CSL, it was the domestic midfield players and high-level foreign strikers that performed more dribbling or running inverted to the threaten area from the flanks, rather than the flank players. Furthermore, one tenth of the L/R Wingers were players starting as full-backs, which indicated that there were couples of full backs in the CSL who could make essential contribution to attacking, but the ratio was relatively small.

*Central Midfielders*

Midfielders are the primary link of a team when building attacks (Clemente et al., 2015), Left/Right Defensive Midfielders were mainly focused on regaining possession and reorganizing from the backfield. In that case, they were reasonable to have a relatively low rating since most of the rating systems tend to fancy on goal scorers (Decroos et al., 2019; Pappalardo et al., 2019). Playmakers are the midfielders who are responsible for providing passes and build up attacking in the center of the pitch. Although acted as the commander on the field, their W/L was the lowest, despite having a second-best rating. Moreover, it is worthwhile to notice that Wide Midfielders performed best among the 4 playing styles which may indicate that owning Midfielders with almighty ability and flank-preferred acting area were potential keys to win in CSL. These met the research results of Wu et al. (2020) that instead of sticking in the middle, winning teams always make a better use of the width of the pitch.

According to studies related to football passing network analysis, midfield players were always the preeminent players in connecting teammates by passing (Clemente et al., 2016a; Clemente et al., 2016b, Yu et al., 2020). While this study further notes that their preferred passing area might be different according to the playing style. Moreover, most of the Central Midfielders were domestic players who simply focused on organizing (Playmakers) and midfield defending (Defensive Midfielders), such tactical role is similar to that of their counterparts Spanish La Liga top teams (Gai et al., 2019; Liu et al., 2016). A possible reason is that most of the powerful foreign midfielders in the CSL were assigned into other positions such as Second Strikers and Mobile Strikers, so that they could offer more direct helps in certain areas, letting their domestic teammates to locate in the middle of the pitch and to center on organizing and defending. Such finding was also supported by the evidence that in the CSL, foreign midfielders outperformed on both goals and assists (Gai et al., 2019). In contrast, strikers in La Liga might strong enough on goal scoring, thus freeing the midfielders to focus on organizing, rather than making threats to the goal (Gonçalves et al., 2014).

*Left/Right Full Backs*

Most of the Full Backs were Wing Backs who started as Left or Right Backs and intended to fulfill the attacking and defensive duties on the flank side. Another 700 more players played as L/R Backs who focused primarily on their defensive duties which explained their lower ratings. Particularly, Wing backs' attacking preferences (e.g., Crosses) may be suppressed by the pressure from the opposition and had to turn to L/R Backs, which may be the reason of the lower Win-Loss ratio. What must be noted is that the preference for crossing is relatively lower than that of dribbling and passing in the flank sides, which may indicate that most of the CSL Wing backs lacked the desires on breaking out from the back and crossing the ball into the box.

In contrast, previous research has pointed out that Wing backs were essential for winning (at both attacking and defending sides) and should contribute more in the final phase of attacking (Konefał et al., 2015; Wu et al., 2020). Specifically, full backs of Spanish top teams tended to execute more attacking actions like crossing and passing than those of the weak teams (Liu et al., 2016). But in the CSL, full backs, almost all of which were domestic players, were not able to accomplish offensive and defensive tasks at both ends (Gai et al., 2019), since they might not meet the increasing match physical demands if they had participated into attacking (Bush et al., 2015; Crossbarhub, 2020; Zhou et al., 2020). Another possible explanation is the lacking of cutting inside ability in most CSL Wing Forwards, since the former players are supposed to provide their Full back teammates enough space to break out to the front area and cross the ball (Crossbarhub, 2020).

*Central Backs*

Nearly all of the Central backs in the CSL were Ball playing Defenders, and only 683 Central Defender were observed. These were in line with a trend that Central backs are becoming more and more important on a team's attacking build up (Clemente et al., 2015; Korte et al., 2019) and the traditional style one are declining (Ayyagari, 2018). Clearly, CSL teams were trying to catch up the tactical trend by requesting their Central backs to convey long passes. However, the minority Central Defenders had a way better W/L, which may indicate that in soccer leagues with lower technical level like the CSL, it is better to let the Central back players focusing on defending, rather than keep thinking about being a long passer (Zhou et al., 2020).

However, Gai et al. (2019) argued that the CSL central defenders preferred more defensive duties instead of being an additional passing options like the defenders of English Premier League (Bush et al., 2015). While this study may provide a new insight from the perspective of playing style that most of the Central backs were trying to accomplish the passing tasks on the field, but they cannot perfectly fulfill it because of the lacking of passing abilities.

## 4.3  Player comparisons

The use of performance profiling in sport allows for a better understanding of players and teams' performances. On the one hand, it could visualize the match actions in different context; on the other hand, the statistical information generated from the profiling process can be added to justify comparisons and modelling performances (Butterworth et al., 2013).

**Alex Teixeira VS. Wei Shihao**

From the high similarity and the general style radar chart, Teixeira and Wei Shihao did share a lot of common grounds. The differences appeared on the preference for taking long shots, dribble in the center part and sending key passes where Teixeira's weights were much higher than Wei Shihao which can also be proved by the result of style clustering. These indicated that Teixeira was more involved into the team's organization and more like a key player of his team than that of Wei Shihao.

**Wu Lei 2017 VS. Wu Lei 2018**

Although Wu Lei's playing style did not change a lot from 2017 to 2018, the differences in the preferred styles and in some specific components could still show some clues. He showed a higher preference for playing in the center front as a Poacher in 2018 than that of 2017. The increased weight on Close shot and Front header combined with the decreased weight on Left shot and Crosses showed this trend too. This change means that he was getting closer to the goal line that made a bigger chance to beat the goalkeeper, which may be the result of his increased match experiences (Kalén et al., 2019).

**Renato Augusto VS. Jonathan Viera**

Above all, the most obvious difference was that Viera played as a Second Striker in almost all his started games in 2018 while Augusto preferred to play in various styles as a Wing Forward or a Center Midfielder. These indicated that Viera preferred to hide behind the main striker and not only sending key passes, but also trying to threat the goal. Augusto was a midfielder who had well-developed physical fitness to enable himself to cover wide field of the pitch, and such player was exactly that strong teams in the CSL should recruit, especially in an era of increasing physical demands in top leagues (Bush et al., 2015).

### *4.4 Practical application*

The current study would provide practical information to club managers and coaches in the CSL at four different levels. Firstly, at a team tactical level, the framework can provide references for coaches on tactical arrangement and formation setting, they may try to enhance the team's performance and win ratio by combining certain types of players or changing tactics by sending on substitute players in certain styles. Secondly, at a club management level, having more knowledge about players' playing style (e.g., similarities) can be quite helpful in the transfer market in case for wasting money on wrong players. Thirdly, at an individual level, players will get to know more about their careers' development and characteristics about their favorite positions and styles. Fourthly, it will provide more specific information about a player's playing style to broadcast companies, thus enhance the watching experience of fans.

## *4.5 Limitation and future works*

This research only considered on-the-ball actions, other important data information in soccer analysis like tracking data have not been used. These types of data can be important on characterizing players' playing style. On the one hand, Garrido et al. (2021) have confirmed in a recent study that the heatmap drawn from event data has a low correlation with the one drawn from tracking data. On the other hand, off-ball actions like positioning and running are essential for positions like central defenders who do not have many chances to touch the ball. Therefore, combining tracking data in the future will be necessary and helpful for making a more precise characterization of the playing style. Furthermore, this study only considered the playing style of outfield players but excluded goalkeepers. Since the style of goalkeeper is more about saving preference and passing only in the back, which the *Player Vectors* model could not attach, it is necessary to develop a data-driven framework to characterize the playing style of goalkeepers in the future.

## 5. Conclusion

This work characterized and analysed the playing style of CSL players in season 2016-2019 based on the state-of-the-art *Player Vectors* framework. The playing styles of the CSL forwards and midfielders were overall match the trend that winning teams performed better in the flank and players who tend and have the ability to cover as much of the field area as possible should be attached more importance to the winning of the match. In contrast, Full backs and Central backs are positions whose playing styles should be strengthened and reconsidered. Meanwhile, position and style clustering results proved that in CSL, better players, especially foreign players are becoming more multifunctional, which was also a widely accepted trend. Furthermore, the player comparison provided and verified the possibility of further using a data-driven model like the *Player Vectors* to analyse and compare the playing styles in certain scenarios.

# Acknowledgement

This is an original manuscript of an article published by Taylor & Francis in Journal of Sports Sciences on 06 Jul 2022, available online: http://www.tandfonline.com/10.1080/02640414.2022.2096771.

# References

Aalbers, B. & Van Haaren, J. (2019). Distinguishing Between Roles of Football Players in Play-by-Play Match Event Data. In U. Brefeld, J. Davis, J. V. Haaren & A. Zimmermann (Eds.), *Machine Learning and Data Mining for Sports Analytics MLSA 2018. Lecture Notes in Computer Science*, vol 11330. (pp. 31-41). Springer, Cham. https://doi.org/10.1007/978-3-030-17274-9_3

Ayyagari, S. (2018, November 21). Why Are Traditional Centre-Backs Diminishing In World Football. Retrieved December 20, 2021, from https://blog.playo.co/why-are-traditional-centre-backs-diminishing/

Bialkowski, A., Lucey, P., Carr, P., Yue, Y., Sridharan, S. & Matthews, I. (2014). *Large-scale analysis of soccer matches using spatiotemporal tracking data.* Paper presented at the 2014 IEEE International Conference on Data Mining, Shenzhen, China.

Bush, M., Barnes, C., Archer, D. T., Hogg, B. & Bradley, P. S. (2015). Evolution of match performance parameters for various playing positions in the English Premier League. *Human Movement Science, 39*, 1-11. https://doi.org/https://doi.org/10.1016/j.humov.2014.10.003

Butterworth, A., O'Donoghue, P. & Cropley, B. (2013). Performance profiling in sports coaching: a review. *International Journal of Performance Analysis in Sport, 13*(3), 572-593. https://doi.org/10.1080/24748668.2013.11868672

Castellano, J. & Pic, M. (2019). Identification and Preference of Game Styles in LaLiga Associated with Match Outcomes. *International Journal of Environmental Research and Public Health, 16*(24), 5090. https://doi.org/10.3390/ijerph16245090

Clemente, F. M., Martins, F. M. L., Wong, P. D., Kalamaras, D. & Mendes, R. S. (2015). Midfielder as the prominent participant in the building attack: A network analysis of national teams in FIFA World Cup 2014. *International Journal of Performance Analysis in Sport, 15*(2), 704-722. https://doi.org/10.1080/24748668.2015.11868825

Clemente, F. M., José, F., Oliveira, N., Martins, F. M. L., Mendes, R. S., Figueiredo, A. J., . . . Kalamaras, D. (2016a). Network structure and centralization tendencies in professional football teams from Spanish La Liga and English Premier Leagues. *Journal of Human Sport and Exercise, 11*(3), 376-389. https://doi.org/10.14198/jhse.2016.113.06

Clemente, F. M., Silva, F., Martins, F. M. L., Kalamaras, D. & Mendes, R. S. (2016b). Performance Analysis Tool for network analysis on team sports: A case study of FIFA Soccer World Cup 2014. *230*(3), 158-170. https://doi.org/10.1177/1754337115597335

Crossbarhub. (2020, May 23). Midfielder Types in Football. Retrieved December 21, 2021, from https://crossbarhub.com/tactical-analysis/midfielder-types-in-football/

Crossbarhub. (2020, October 13). Types of Fullbacks. Retrieved December 21, 2021, from https://crossbarhub.com/tactical-analysis/types-of-fullbacks/

Decroos, T., Bransen, L., Van Haaren, J. & Davis, J. (2019). *Actions Speak Louder than Goals : Valuing Player Actions in Soccer*. Paper presented at the Proceedings of the 25th ACM SIGKDD International Conference on Knowledge Discovery & Data Mining. Anchorage, AK, USA. https://doi.org/ 10.1145/3292500.3330758

Decroos T., Davis J. (2020) Player Vectors: Characterizing Soccer Players' Playing Style from Match Event Streams. In: Brefeld U., Fromont E., Hotho A., Knobbe A., Maathuis M., Robardet C. (Eds) *Machine Learning and Knowledge Discovery in Databases. ECML PKDD 2019. Lecture Notes in Computer Science*, vol 11908 (pp. 569-584). Springer, Cham. https://doi.org/10.1007/978-3-030-46133-1_34

Duarte, R., Araújo, D., Correia, V. & Davids, K. (2012). Sports Teams as Superorganisms. *Sports Medicine, 42*(8), 633-642. https://doi.org/10.1007/BF03262285

Fernandez-Navarro, J., Fradua, L., Zubillaga, A., Ford, P. R. & McRobert, A. P. (2016). Attacking and defensive styles of play in soccer: analysis of Spanish and English elite teams. *Journal of Sports Sciences, 34*(24), 2195-2204.

https://doi.org/10.1080/02640414.2016.1169309

Gai, Y., Volossovitch, A., Lago, C. & Gómez, M.-Á. (2019). Technical and tactical performance differences according to player's nationality and playing position in the Chinese Football Super League. *International Journal of Performance Analysis in Sport, 19*(4), 632-645. https://doi.org/10.1080/24748668.2019.1644804

Garcia-Aliaga, A., Marquina, M., Coteron, J., Rodriguez-Gonzalez, A. & Luengo-Sanchez, S. (2020). In-game behaviour analysis of football players using machine learning techniques based on player statistics. *International Journal of Sports Science & Coaching, 16*(1), 148-157. https://doi.org/10.1177/1747954120959762

Garrido, D., Burriel, B., Resta, R., del Campo, R. L. & Buldu, J. (2021). Heatmaps in soccer: event vs tracking datasets. *arXiv preprint*. https://arxiv.org/abs/2106.04558

Goes, F., Kempe, M., van Norel, J. & Lemmink, K. A. P. M. (2021). Modelling team performance in soccer using tactical features derived from position tracking data. *IMA Journal of Management Mathematics, 32*(4), 519–533. https://doi.org/10.1093/imaman/dpab006

Goes, F., Meerhoff, L. A., Bueno, M. J. O., Rodrigues, D. M., Moura, F. A., Brink, M. S., . . . Lemmink, K. (2020). Unlocking the potential of big data to support tactical performance analysis in professional soccer: A systematic review. *European Journal of Sport Science, 21*(4), 481-496. https://doi.org/10.1080/17461391.2020.1747552

Gómez, M.-Á., Mitrotasios, M., Armatas, V. & Lago-Penas, C. (2018). Analysis of playing styles according to team quality and match location in Greek professional soccer. *International Journal of Performance Analysis in Sport, 18*(6), 986-997. https://doi.org/10.1080/24748668.2018.1539382

Gonçalves, B. V., Figueira, B. E., Maçãs, V. & Sampaio, J. (2014). Effect of player position on movement behaviour, physical and physiological performances during an 11-a-side football game. *Journal of Sports Sciences, 32*(2), 191-199.

https://doi.org/10.1080/02640414.2013.816761

Gudmundsson, J. & Horton, M. (2017). Spatio-Temporal Analysis of Team Sports. *ACM Computing Surveys, 50*(2), 1-34. https://doi.org/10.1145/3054132

Guide to FM. (2020). Central Attack Roles. Retrieved December 26, 2021, from https://www.guidetofm.com/tactics/central-attack-roles/

Hewitt, A., Greenham, G. & Norton, K. (2016). Game style in soccer: what is it and can we quantify it? *International Journal of Performance Analysis in Sport, 16*(1), 355-372. https://doi.org/10.1080/24748668.2016.11868892

Kalén, A., Rey, E., de Rellán-Guerra, A. S. & Lago-Peñas, C. (2019). Are Soccer Players Older Now Than Before? Aging Trends and Market Value in the Last Three Decades of the UEFA Champions League. *Frontiers in Psychology, 10*. https://doi.org/10.3389/fpsyg.2019.00076

Konefał, M., Chmura, P., Andrzejewski, M., Pukszta, D. & Chmura, J. (2015). Analysis of match performance of full-backs from selected European soccer leagues. *Central European Journal of Sport Sciences Medicine, 11*(3), 45-53. https://doi.org/10.18276/cej.2015.3-05

Korte, F., Link, D., Groll, J. & Lames, M. (2019). Play-by-Play Network Analysis in Football. *Frontiers in Psychology, 10(1738)*. https://doi.org/10.3389/fpsyg.2019.01738

Lago-Peñas, C., Gómez-Ruano, M. & Yang, G. (2018). Styles of play in professional soccer: an approach of the Chinese Soccer Super League. *International Journal of Performance Analysis in Sport, 17*(6), 1073-1084. https://doi.org/10.1080/24748668.2018.1431857

Lasek, J., Szlávik, Z. & Bhulai, S. (2013). The predictive power of ranking systems in association football. *International Journal of Applied Pattern Recognition, 1*(1), 27-46. https://doi.org/10.1504/IJAPR.2013.052339

Lee, D. D. & Seung, H. S. (2000). *Algorithms for non-negative matrix factorization*. Paper presented at the Proceedings of the 13th International Conference on Neural Information Processing Systems, Denver, CO, USA.

Li, Y., Ma, R., Gonçalves, B., Gong, B., Cui, Y. & Shen, Y. (2020). Data-driven team

ranking and match performance analysis in Chinese Football Super League. *Chaos, Solitons & Fractals, 141*, 110330. https://doi.org/10.1016/j.chaos.2020.110330

Liu, H., Gómez, M.-Á., Gonçalves, B. & Sampaio, J. (2016). Technical performance and match-to-match variation in elite football teams. *Journal of Sports Sciences, 34*(6), 509-518. https://doi.org/10.1080/02640414.2015.1117121

Liu, H., Hopkins, W., Gómez, M.-Á. & Molinuevo, S. J. (2013). Inter-operator reliability of live football match statistics from OPTA Sportsdata. *International Journal of Performance Analysis in Sport, 13*(3), 803-821. https://doi.org/10.1080/24748668.2013.11868690

Lord, F., Pyne, D. B., Welvaert, M. & Mara, J. K. (2020). Methods of performance analysis in team invasion sports: A systematic review. *Journal of Sports Sciences, 38*(20), 2338-2349. https://doi.org/10.1080/02640414.2020.1785185

Mazurek, J. (2018). Which football player bears most resemblance to Messi? A statistical analysis. *arXiv preprint*. https://arxiv.org/abs/1802.00967

McHale, I. G., Scarf, P. A. & Folker, D. E. (2012). On the Development of a Soccer Player Performance Rating System for the English Premier League. *Interfaces, 42*(4), 339-351. https://doi.org/10.1287/inte.1110.0589

McLean, S., Salmon, P. M., Gorman, A. D., Naughton, M. & Solomon, C. (2017). Do inter-continental playing styles exist? Using social network analysis to compare goals from the 2016 EURO and COPA football tournaments knock-out stages. *Theoretical Issues in Ergonomics Science, 18*(4), 370-383. https://doi.org/10.1080/1463922x.2017.1290158

Pappalardo, L., Cintia, P., Ferragina, P., Massucco, E., Pedreschi, D. & Giannotti, F. (2019). PlayeRank: Data-driven Performance Evaluation and Player Ranking in Soccer via a Machine Learning Approach. *ACM Transactions on Intelligent Systems and Technology, 10*(5), 1-27. https://doi.org/10.1145/3343172

Pedregosa, F., Varoquaux, G., Gramfort, A., Michel, V., Thirion, B., Grisel, O., . . . Dubourg, V. (2011). Scikit-learn: Machine learning in Python. *the Journal of machine Learning research, 12*(85), 2825-2830.

Peña, J. L. & Navarro, R. S. (2015). Who can replace Xavi? A passing motif analysis of football players. *arXiv preprint.* https://arxiv.org/abs/1506.07768

Rousseeuw, P. J. (1987). Silhouettes: A graphical aid to the interpretation and validation of cluster analysis. *Journal of Computational and Applied Mathematics, 20*, 53-65. https://doi.org/ 10.1016/0377-0427(87)90125-7

Whoscored. (2018). Super League Player Statistics 2018. Retrieved December 26, 2021, from https://www.whoscored.com/Regions/45/Tournaments/162/Seasons/7242/Stages/15995/PlayerStatistics/China-Super-league-2018

Wu, Y., Xia, Z., Wu, T., Yi, Q., Yu, R. & Wang, J. (2020). Characteristics and optimization of core local network: Big data analysis of football matches. *Chaos, Solitons & Fractals, 138,* 110136. https://doi.org/ 10.1016/j.chaos.2020.110136

Yu, Q., Gai, Y., Gong, B., Gómez, M.-Á. & Cui, Y. (2020). Using passing network measures to determine the performance difference between foreign and domestic outfielder players in Chinese Football Super League. *International Journal of Sports Science & Coaching, 15*(3), 398-404. https://doi.org/10.1177/1747954120905726

Zhou, C., Calvo, A. L., Robertson, S. & Gómez, M.-Á. (2020). Long-term influence of technical, physical performance indicators and situational variables on match outcome in male professional Chinese soccer. *Journal of Sports Sciences, 39*(6), 598-608. https://doi.org/10.1080/02640414.2020.1836793

Zhou, C., Lago-Peñas, C., Lorenzo, A. & Gómez, M.-Á. (2021). Long-Term Trend Analysis of Playing Styles in the Chinese Soccer Super League. *Journal of Human Kinetics, 79*(1), 237-247. https://doi.org/10.2478/hukin-2021-0077

# Supplementary materials

**Supplementary Table 1.** The definitions and available results and attributes of all the events in the data-set

| Event name | Definitions | Event results | Event attributes |
|---|---|---|---|
| Simple pass | Any intentional played ball from one player to another. | | |
| Long pass | A lofted ball where there is a clear intended recipient, must be over shoulder height and using the passes height to avoid opposition players or a long high ball into space or into an area for players to chase or challenge for the ball | | |
| Throw in | Throw in | | |
| Goal kick | Goal kick | Assist，Keypass, Successful, Unsuccessful | Through、simple |
| Freekick pass | A pass made from a freekick | | |
| Cross | Any intentional played ball from a wide position intending to reach a team mate in a specific area in front of the goal. | | |
| Freekick cross | A cross made from a freekick | | |
| Corner cross | A cross made from a corner | | |
| Shot | An attempt to score a goal, made with any (legal) part of the body, either on or off target | | |
| Freekick shot | Any attempts created directly from the free kick itself (unassisted). | Goal, On target, Off target, On post, Blocked | Out box, In box |
| Penalty shot | A shot made from a penalty | | |
| Corner shot | A shot made by the corner taker | | |
| Chance missed | A big chance opportunity when the player does not get a shot away, typically given for big chance attempts where the player shooting completely misses the ball (air shot) but can also be given when the player has a big chance opportunity to shoot and decides not to, resulting in no attempt occurring in that attack. | Unsuccessful | Simple |
| Offside | Awarded to the player deemed to be in an offside position where a free kick is awarded | Unsuccessful | Simple |
| Own goal | An own goal is usually awarded if the attempt is off target and deflected into the goal by an opponent. | Own goal | Simple |
| Clearance | A defensive action where a player kicks the ball away from his own goal with no intended recipient. | Successful, Unsuccessful | Simple |
| Interception | where a player reads an opponent's pass and intercepts the ball by moving into the line of the intended pass. | Successful | Simple |
| Tackle | where a player connects with the ball in a ground challenge where he successfully takes the ball away from the player in possession. | Successful, Unsuccessful | Simple |
| Defensive duel | Player has been beaten in one-on-on | Unsuccessful | Simple |
| Blocked pass | When a player tries to cut out an opposition pass by any means. Similar to an interception except there is much less reading of the pass. | Successful | Simple |
| Blocked shot | A player blocks a shot on target from an opposing player. | Successful | Simple |
| Defensive foul | Any infringement that is penalised as foul play by a referee in defending. | Unsuccessful, Red card, Second yellow card, Yellow card | Simple |
| Offensive foul | Any infringement that is penalised as foul play by a referee in offending. | | Simple |
| Red card | | Unsuccessful | Simple |
| Second yellow card | | | Simple |
| Yellow card | | | Simple |
| Aerial duel | This is where two players challenge in the air against each other. | Successful, Unsuccessful | Simple |
| Shield ball out | Where a player shields the ball from an opponent and is successful in letting it run out of play. Can be offensive (to win a corner or throw in up field) or defensive (winning a throw in or goal kick). | Successful | Simple |
| Ball recovery | where a player recovers the ball in a situation where neither team has possession or where the ball has been played directly to him by an opponent, thus securing possession for their team. | Successful | Simple |
| Offside provoked | The deepest player in the defensive line when an offside has been given. | Unsuccessful | Simple |
| Foul conceded | any infringement that is penalised as foul play by a referee. | Successful | Simple |
| TakeOn | an attempt by a player to beat an opponent when they have possession of the ball | Successful, Unsuccessful | Simple |
| Dribble | Player dribbles at least 3 meters with the ball | Successful | Simple |
| Touch | When the ball bounces off a player and there is no intentional pass, | Successful, Unsuccessful | Simple |
| Dispossessed | A player is in possession but not attempting to "beat" his opponent and get tackled | Unsuccessful | Simple |
| Error | When a player makes an error, which leads to a goal or shot conceded. Also used for spills and attempted claims or saves by a goalkeeper which directly leads to a second attempt to score. | Unsuccessful | Simple |
| Keeper save | A goalkeeper preventing the ball from entering the goal with any part of his body when facing an intentional attempt from an opposition player. | Successful | Simple |
| Keeper pickup | When the goal keeper picks up the ball and his side regain possession, similar to recovery, however, the goal keeper picks the ball up. | Successful | Simple |
| Penalty faced | A goalkeeper is facing a penalty | Successful, Unsuccessful | Simple |
| Keeper sweeper | Anytime a goalkeeper anticipates danger and rushes off their line to try to either cut out an attacking pass (in a race with the opposition player) or to close-down an opposition player. | Successful, Unsuccessful | Simple |
| Keeper smother | A goalkeeper who comes out and claims the ball at the feet of a forward gets a smother, similar to a tackle, however, the keeper must hold onto the ball to award a smother. | Successful | Simple |

| Keeper punch | A high ball that is punched clear by the goalkeeper. The keeper must have a clenched fist and attempting to clear the high ball rather than claim it. | Successful | Simple |
| --- | --- | --- | --- |
| Keeper claim | A high ball played into the penalty area that is caught by the goalkeeper. | Successful | Simple |
| CrossNotClaimed | When a goalkeeper comes off his goal line to claim a high ball (attempting a catch) and misses the ball. | Unsuccessful | Simple |

**Supplementary Table 2.** The definition and number of appearances of each style in each position

| Position | Style | Appearances | | | | Definitions |
|---|---|---|---|---|---|---|
| | | Total | Start Position | Domestic | Foreign | |
| ST | Second Striker | 1394 | MF 508 | 176 | 332 | Strikers or attacking midfielders who tend to move deep into the midfield to help team's organizing |
| | | | FW 695 | 101 | 594 | |
| | Target Man | 570 | FW 533 | 102 | 431 | Strikers who need to open up spaces for key players by their physicality and aerial presence |
| | Mobile Striker | 646 | LM 119 | 72 | 47 | Strikers who like to play on the flanks and cutting into the middle |
| | | | RM 104 | 59 | 45 | |
| | | | FW 245 | 59 | 186 | |
| | | | MF 177 | 52 | 125 | |
| | Poacher | 613 | FW 523 | 113 | 410 | Strikers focusing on putting the ball in the back of the net |
| CM | L/R Defensive Midfielder | 1563 | MF 1427 | 1097 | 330 | Midfielders who are good at tackling, interception and physical duels |
| | Playmaker | 1549 | MF 1403 | 989 | 414 | Midfielders who undertake the main organizing task of the team |

| | | | | | | | |
|---|---|---|---|---|---|---|---|
| | Wide Midfielder | 544 | MF | 317 | 225 | 92 | Midfielders who have a large scope of running area and good at playing on the flanks |
| L/RW | | | L/RM | 1523 | 1126 | 397 | |
| | L/R Winger | 2531 | L/RB | 230 | 226 | 4 | Players who are good at breaking the defensive line and making shots or crosses on the flanks |
| | | | MF | 481 | 355 | 126 | |
| | | | FW | 115 | 44 | 71 | |
| | | | LM | 126 | 63 | 63 | Players who tend to build up attacks on the flanks or carry the ball to the middle of the field |
| | Inside Forward | 890 | RM | 124 | 66 | 58 | |
| | | | FW | 161 | 36 | 125 | |
| | | | L/RM | 311 | 297 | 14 | Full backs who tend to break out from the back and making crosses or passes |
| FB | L/R Wing Back | 3252 | L/RB | 2599 | 2578 | 21 | |
| | | | MF | 252 | 249 | 3 | |
| | L/R Back | 733 | L/RB | 283 | 283 | 0 | Full backs who tend to focus on defending |
| | | | DF | 239 | 201 | 38 | |
| CB | L/R Ball Playing Defender | 3729 | DF | 3279 | 2411 | 868 | Central backs who undertake the backfield carrying and passing of the team |
| | | | MF | 278 | 241 | 37 | |
| | Central Defender | 685 | DF | 611 | 449 | 162 | Central backs who tend to focus on defending |

Notes: **ST** = Strikers, **L/RW** = Left/Right Wing forwards, **CM** = Central Midfielders, **L/RFB** = Left/Right Full Backs, **CB** = Central Backs, **FW** = Forward, **MF** = Midfielder, **CD**=Center Defender, **L/RM** = Left/Right Midfielder, **L/RD** = Left/Right Defender